\documentclass[letterpaper, 10pt, conference]{ieeeconf}
\usepackage[T1]{fontenc}
\usepackage[utf8]{inputenc}
\usepackage{textcomp}
\usepackage{graphicx}
\usepackage{amsmath,amssymb}
\usepackage{booktabs}
\usepackage{array}
\usepackage{url}
\graphicspath{{figures/}}
\makeatletter
\def\bstctlcite{\@ifnextchar[{\@bstctlcite}{\@bstctlcite[@auxout]}}
\def\@bstctlcite[#1]#2{\@bsphack
  \@for\@citeb:=#2\do{%
    \edef\@citeb{\expandafter\@firstofone\@citeb}%
    \if@filesw\immediate\write\csname #1\endcsname{\string\citation{\@citeb}}\fi}%
  \@esphack}
\makeatother

\title{\LARGE \bf Confidently Lost: Silent Failure and Edge Deployment\\of Dense Visual SLAM under Photon Starvation}
\author{Shrikrishna Bhagirath Rajule, Srinath Neelamagam, and Aniket Bera\\
{\normalsize Purdue University, West Lafayette, IN 47907, USA}}

\begin{document}
\maketitle
\bstctlcite{IEEEexample:BSTcontrol}
\thispagestyle{empty}
\pagestyle{empty}

\begin{abstract}
Dense visual SLAM built on learned two-view geometry is evaluated in good light; in low light it fails \emph{silently}. Two frames of 9-level quantized noise, no scene at all, give MASt3R-SLAM's network a cross-view geometric self-consistency of 0.936; a real dark frame pair gives 0.935. Starved of structure, the network invents the same geometry for both views and the system trusts the agreement: on a dark sequence it consumes 1\,605 frames, emits 7 keyframes, rejects no factor-graph edges and reports full frame rate, while over 95\% of the correspondences it accepts violate the ground-truth epipolar geometry. One inflated statistic is read by two subsystems against thresholds in opposite directions, so keyframing starves while relocalization accepts almost every candidate. The same statistic, read at keyframe-pair baselines, separates the confident collapse from working runs on the ETH3D dark sequences without ground truth or extra computation (in-sample). Low-light-adapted weights take tracked trajectories from 1/6 to 5/6, as does retuning two thresholds; neither yields accurate ones, because the adaptation withdraws confidence from wrong geometry rather than finding right geometry. Replayed on a Jetson AGX Orin carried by a legged robot, the mechanism's signature reproduces on the two darkest of four routes, and which of its faces appears varies with route geometry and sensor encoding. The network does not lose confidence in the dark; it manufactures it.
\end{abstract}

\section{Introduction}\label{sec:intro}

\begin{figure*}[t]
\centering
\includegraphics[width=0.56\textwidth]{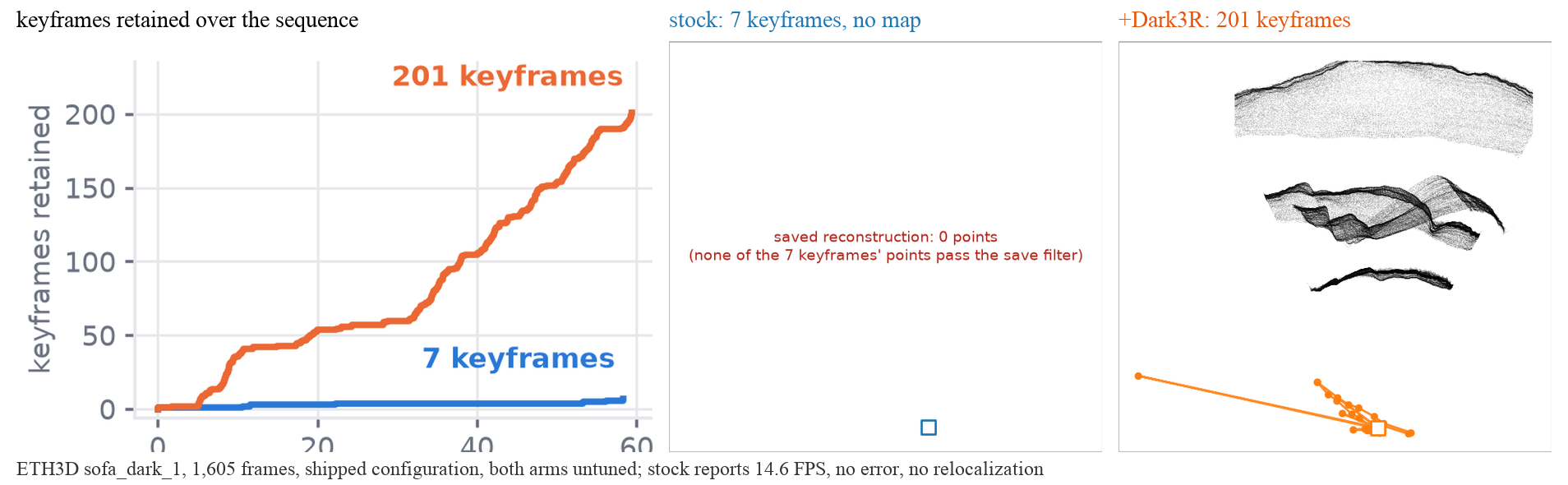}
\caption{Silent failure. ETH3D \texttt{sofa\_dark\_1}, 1\,605 frames, shipped configuration, both arms untuned. Left: keyframes retained over the sequence, 7 for stock against 201 for the adapted arm. Middle and right: the saved reconstruction of each run, top-down, with its keyframe poses (square: start). Stock reported 14.6~FPS, no error, and no relocalization, and saved an empty reconstruction.}
\label{fig:hook}
\vspace{-3.5mm}
\end{figure*}

A robot inspecting a tunnel at night needs a SLAM system that runs on the computer it carries, in the light it has. Dense visual SLAM built on learned two-view geometry is the strongest candidate: MASt3R-SLAM~\cite{murai2025mast3rslam} regresses dense pointmaps and descriptors from a foundation model and reconstructs without calibration in real time. But every quantitative result for this class of system is measured in good light, on the edge as throughput alone~\cite{ec3r2026}, or on the pairwise network alone under synthetic darkening~\cite{kumar2026uavslam}; whether it can be deployed in the dark on an edge computer was untested. This paper asks what happens when it is: the specific form its failure takes under photon starvation, the mechanism, the health signal that exposes the failure at no cost, what the available remedy buys and costs, and which of these carry to a Jetson AGX Orin on a legged robot.

The obvious remedy exists: Dark3R~\cite{dark3r2026} adapts the same MASt3R backbone to photon-starved raw imagery by LoRA distillation~\cite{hu2022lora}, offline, where every image is available and geometry is globally optimized. SLAM is causal and must commit to a keyframe, a match and a pose now; whether a distilled low-light prior survives that transition was untested.

\textbf{We find that it survives as behavior but not as geometry, and that the more interesting result is what happens to the system it was meant to fix.} Loading Dark3R's weights into MASt3R-SLAM, changing nothing else, takes usable trajectories on the ETH3D-SLAM dark sequences~\cite{schoeps2019eth3d} from 1 of 6 to 5 of 6, \emph{usable} being a workload criterion ($\geq 50$ keyframes, a converging alignment), not an accuracy one: every dark trajectory has meter-scale error, and over 95\% of the correspondences either model accepts violate the ground-truth epipolar geometry (Sec.~\ref{sec:mech}). Retuning two of the stock system's gate thresholds also reaches 5 of 6 (Sec.~\ref{sec:main}), with the same wrong correspondences; no gate retuning leaves the degenerate regime at the shipped acceptance radius, and adaptation does. On \texttt{sofa\_dark\_1} stock MASt3R-SLAM consumes 1\,605 frames, emits 7 keyframes and reports 14.6~FPS, and every signal a deployment could monitor reads nominal: 0 skipped frames, 0 Cholesky failures, 0 relocalization attempts, no factor-graph edge rejected, a trajectory file the standard evaluation scores. It leaves a seven-keyframe map of a 1\,605-frame walk (Fig.~\ref{fig:hook}), silent to the system, to the evaluation protocol (an ATE over the keyframe file cannot tell a seven-keyframe map from a complete one) and to any downstream consumer.

The mechanism is a property of the network's output at the matcher's acceptance radius, not of any gate threshold. MASt3R-SLAM validates a correspondence by \emph{geometric self-consistency}: a pixel carried through the network's cross-view pointmap must land within 10~cm of the other view's own pointmap. In near-darkness the two predictions become spuriously consistent, the network emits geometry that agrees with itself about a scene it cannot see, and validity reaches 0.93. The keyframe rule fires when its confident-match operand falls \emph{below} 0.333; on the silent runs that operand's median is 0.85--0.89, so the rule fires six times in 1\,605 frames. Relocalization accepts a candidate when the same quantity rises \emph{above} 0.3: on \texttt{plant\_dark} it accepts 497 of 543 attempts, rejecting 21.5\% of the edges they propose. One cause, two opposite symptoms, which we call the \emph{silent} exit (starved keyframing at nominal FPS) and the \emph{loud} exit (thrashing: unbounded relocalization onto wrong anchors).

The diagnosis yields a detector that costs no computation. Geometric match validity at keyframe-pair baselines, $g$, is computed at every factor-graph edge, and across a 24-run grid on 8-bit ETH3D it separates the confident collapse from tracking operation with no overlap (Sec.~\ref{sec:mech}). A spuriously \emph{high} match rate is the alarm. It is a property of the system's state, not of the input: on the same frames stock reads 0.929 and the adapted model 0.371. It does not certify the map that remains, and its healthy level is a property of the camera.

Our contributions are: (1)~\textbf{a mechanism}, verified on held-out sequences and by per-gate intervention, for why dense learned SLAM fails silently in low light: the network does not lose confidence in the dark, it manufactures it, and the same number that starves keyframing makes relocalization accept everything; (2)~\textbf{a detector that needs no ground truth and no extra computation}, with an operating point and its error rates over 66 labeled runs; (3)~\textbf{a controlled comparison of remedies}, retuning, an enhancement front-end, weight adaptation and a training-free dose between stock and adapted weights, in which, at the shipped acceptance radius, only adaptation leaves the degenerate regime, and a partial dose removes its one measured good-light degeneracy while its ATE cost scales with the dose; (4)~\textbf{a scope test on a robot}: on a Jetson AGX Orin (memory law, Orin NX ceiling, flat power) and a quadruped's 8-bit camera over four routes, the mechanism's signature reproduces on the two darkest routes, the adaptation covers a level stock loses on two of three turn-free routes, and which face of the fault appears varies with route geometry and sensor encoding.

\section{Related work}\label{sec:related}

\textbf{Learned dense geometry and its monitoring.} DUSt3R \cite{wang2024dust3r} recast stereo as pointmap regression, MASt3R~\cite{leroy2024mast3r} added a descriptor head, and MASt3R-SLAM~\cite{murai2025mast3rslam} embedded the model in a causal pipeline; feed-forward successors~\cite{wang2025vggt,yang2025pi3} and the SLAM systems on them~\cite{maggio2025vggtslam} are evaluated in good light. Learned trackers under corruption trade catastrophic loss for sustained drift~\cite{thomas2026failuredrift}; a low-light report on six trackers~\cite{basystyi2026lowlightslam} includes no pointmap system. We build on MASt3R-SLAM because the adaptation is a LoRA on cross-attention projections VGGT-family architectures lack. Run-time monitoring learns error predictors~\cite{rabiee2020ivslam} or risk scores from residuals~\cite{gaus2025super}, and the \emph{SLAM confidence trap}~\cite{sansoni2025confidencetrap} argues the field ships precise systems with no consistent uncertainty; ours is an unlearned statistic the system already computes, for a failure none describes: a system that reports nothing wrong. LeanGate~\cite{xiong2026leangate} skips frames a predictor deems redundant; Sec.~\ref{sec:mech} describes a regime in which apparent redundancy is the failure.

\textbf{Low light, enhancement, other sensors, embedded compute.} Dark3R~\cite{dark3r2026} distills MASt3R on real photon-starved raw for offline structure from motion; DarkVGGT~\cite{darkvggt2026} adapts VGGT with LoRA but requires a thermal camera at inference, and names the \emph{daylight tax} we measure inside SLAM. Twilight SLAM~\cite{twilightslam2023} found enhancement a mixed remedy for feature-based pipelines; DarkIR~\cite{darkir2025} degrades dense SLAM (Sec.~\ref{sec:enh}). Near-infrared illumination~\cite{kong2021nir,clearir2026} and thermal fusion~\cite{darkvggt2026} add a sensor; we ask how far a passive camera can be pushed before they are needed. EC3R-SLAM~\cite{ec3r2026} measures MASt3R-SLAM at 0.1~FPS on an Orin NX; we do not compete on throughput. The closest evaluation~\cite{kumar2026uavslam} reports MASt3R at 93\% tracking success under synthetic low light on Jetson hardware, but runs the pairwise network without the keyframe and relocalization subsystems that fail here, under a 2~m, 30$^\circ$ criterion blind to a four-keyframe map.

\section{Method}\label{sec:method}

\subsection{Where confidence enters MASt3R-SLAM}\label{sec:prelim}

For a keyframe pair $(i,j)$ the backbone produces four pointmaps in one pass, $X^{ii}$, $X^{jj}$ and the cross-view $X^{ji}$, $X^{ij}$, with per-pixel confidences; the system forms a symmetric matching confidence $Q_{i\to j}=\sqrt{Q^{ii}[\mathrm{idx}_{i\to j}]\,Q^{ji}}$ gated against $\tau_Q$ (\texttt{Q\_conf}, shipped at 1.5). \textbf{Geometric match validity} $g$ is the fraction of pixels whose correspondence is self-consistent: each pixel of view $j$ is carried through $X^{ji}$, projected onto view $i$'s rays by Gauss--Newton, and accepted if the projection converges inside the image and lands within \texttt{dist\_thresh} $=0.1$ of $X^{ii}$ at that pixel. Descriptors only refine the pixel afterwards; neither they nor $Q$ change validity, so $g$ is a statistic of the network's two geometry predictions agreeing with each other, upstream of every gate. \textbf{Match fraction} $m$, the post-gate statistic, is $g$'s mask and $Q>\tau_Q$: valid \emph{and} confident. The tracker's keyframe test reads the per-frame form of $m$ with the fraction of distinct keyframe pixels reached, $\min(m_k,u_f)$, and creates a keyframe when it falls below \texttt{match\_frac\_thresh} $=0.333$; relocalization accepts a candidate when $m$ rises above \texttt{reloc.min\_match\_frac} $=0.300$. \textbf{These rules point in opposite directions against near-identical thresholds}, benign in normal light and the root of both failure modes we observe.

\subsection{Adapting the backbone}\label{sec:adapt}

Dark3R supplies LoRA adapters trained on photon-starved raw. We merge them into the base weights offline, producing a drop-in checkpoint with stock's key set (1\,017 tensors) and no change to the SLAM code: the \emph{adapted} arm, \textbf{+Dark3R} in tables and \texttt{merged} in run names. Exactly 288 tensors differ, 96 each in the encoder, decoder and second decoder blocks, and the prediction heads are bit-identical (Dark3R's figure describes fine-tuning the heads; the shipped checkpoint leaves them untouched). The adaptation does not change matching alone: on 24 identical-input frame pairs per sequence the median relative pointmap change between arms is 168\% in the dark against 11.5\% in bright while the match fraction moves 32\% (the denominator is stock's own dark pointmap: the direction, not the multiple, is the result). These pairs are a frame against its own keyframe a few frames apart; their short-baseline validity, $g_{\mathrm{short}}$, is 0.930 for stock in the dark, 0.631 after adaptation, 0.89--0.90 for both arms in good light. A short-baseline pair is self-consistent for any model, so $g_{\mathrm{short}}$ is not the detector (Sec.~\ref{sec:mech}); what moves is the dark value. The adaptation's effect is a withdrawal of confidence, not a correction: it removes the cross-view agreement darkness manufactures, but the correspondences it still accepts are as wrong as stock's against ground truth. Less confident, not more correct (Sec.~\ref{sec:ablate} locates the effect in the weights).

\subsection{Dosing the adaptation}\label{sec:dose}

Because the merge folds low-rank updates without rescaling, $\Delta=W_{\mathrm{merged}}-W_{\mathrm{stock}}$ \emph{is} the released adaptation, and $W(s)=W_{\mathrm{stock}}+s\,\Delta$, $s\in[0,1]$, is a straight line from stock to the full merge. We build $W(s)$ once, before the SLAM process starts; the operator is plain weight-space interpolation, and what is new is what the dose does inside a causal SLAM loop.

\subsection{Instrumenting the failure, and a latent defect}\label{sec:instrument}

We add a read-only diagnostic at the factor-graph edge-construction site, enabled by environment variable, recording per call the distribution of $Q$, $g$, $m$ per edge direction, edges offered and rejected, and an explicit caller tag. We also found that \texttt{local\_opt.window\_size} is read and never used: edges accumulate without eviction and, under low light where each relocalization attempt adds edges, exhaust a 15.5~GiB GPU at keyframe 273; we implement the windowing the parameter implies (an edge survives while either endpoint is inside the window). An independent fork found the same defect; we claim the measurement of its cost.

\subsection{Evaluation protocol}\label{sec:protocol}

Under the evaluation configuration the pipeline is bit-exactly deterministic within a fixed hardware configuration (byte-identical trajectories on repeat), so we report one run per configuration; determinism does not extend to timing nor across hardware (a Jetson power-mode change moved the keyframe count on 3 of 6 runs, one by 43\%). Three trajectory protocols appear and are never mixed in one table: ATE over \emph{keyframe} poses against ETH3D ground truth after Sim(3) alignment; per-frame trajectories against COLMAP references for the raw scenes; and keyframe poses against the robot's odometry, a reference rather than ground truth (Sec.~\ref{sec:robot}). Every frame-rate figure carries its keyframe count. $\tau_Q$ was tuned on two sequences (\texttt{einstein\_dark}, \texttt{plant\_dark}) with the other four held out; the gate-threshold retune grid of Sec.~\ref{sec:main} ran on all six.

\section{Experiments}\label{sec:exp}

\textbf{Setup.} ETH3D-SLAM dark sequences (Vicon ground truth, 281--2\,035 frames); 2$\times$ RTX 5080; window 120, single-threaded, calibrated intrinsics; arms \textbf{stock}, \textbf{+DarkIR} (enhance then stock), \textbf{+Dark3R}. A trajectory is \emph{usable} if it has $\geq 50$ keyframes and Sim(3) alignment converges, a workload criterion, not an accuracy one.

\subsection{Main result at the shipped configuration}\label{sec:main}

\begin{table}[t]
\caption{Shipped configuration, both arms untuned, RTX 5080. ATE (m) over keyframe poses after Sim(3) alignment; ``--'' = fewer than 50 keyframes; ``fails'' = alignment does not converge.}
\label{tab:main}
\centering\scriptsize
\setlength{\tabcolsep}{3pt}\scriptsize\begin{tabular}{lrrrrr}
\toprule
Sequence & frames & stock kf & stock ATE & +Dark3R kf & +Dark3R ATE \\
\midrule
\texttt{einstein\_dark} & 2035 & 320 & 1.522 & 358 & 1.639 \\
\texttt{plant\_dark} & 1140 & 506 & fails & 304 & \textbf{0.871} \\
\texttt{kidnap\_dark} & 526 & 15 & -- & 125 & \textbf{1.217} \\
\texttt{sofa\_dark\_1} & 1605 & \textbf{7} & -- & \textbf{201} & \textbf{1.320} \\
\texttt{sofa\_dark\_2} & 281 & 4 & -- & 1 & -- \\
\texttt{sofa\_dark\_3} & 1187 & \textbf{4} & -- & \textbf{160} & \textbf{1.298} \\
\midrule
Usable & & \textbf{1/6} & & \textbf{5/6} & \\
\bottomrule
\end{tabular}
\vspace{-2mm}
\end{table}

Table~\ref{tab:main} is what a practitioner obtains by loading the checkpoint: 1/6 to 5/6, no tuning. The count is not the claim. Stock's usable count is 1/6 at the shipped gate, 2/6 at its own best $\tau_Q$, and 5/6 with both match-fraction thresholds raised to 0.80: three tunings of one system. At 0.80, \emph{stock} weights reach 5/6 with lower ATE than the adapted model on 4 of 6. Across 42 such retuned stock runs at matched protocol, $g$ spans 0.369--0.933 and 41 of 42 stay above the adapted model's ceiling of 0.451 (the exception, \texttt{kidnap\_dark} at 0.369, is stock's established value there), and at 0.80 the retuned silent sequences keep the inflated statistic (post-gate $m$ 0.828 and 0.848 on \texttt{sofa\_dark\_1/3}). Retuning moves the thresholds to meet the inflated statistic; it does not un-inflate it, touches neither the matcher nor the gate's operand, and so accepts exactly the correspondences stock accepts, 97--100\% of which violate the epipolar geometry (Sec.~\ref{sec:mech}); the matcher's acceptance radius is a different lever, treated there. No threshold transfers either: at 0.80 the keyframe rate spans 52--352 per 1\,000 frames and one sequence overflows the store. \texttt{sofa\_dark\_2} (281 frames) yields 4 and 1 poses at the shipped thresholds but $\geq 50$ under retuned stock at 0.80: a limitation of the configuration, not of length. \texttt{einstein\_dark}'s ATE is bistable (1.503 or 1.639 under solver-tolerance changes), so its row carries no ordering.

\subsection{Mechanism and detector}\label{sec:mech}

\begin{figure*}[t]
\centering
\includegraphics[width=0.7\textwidth]{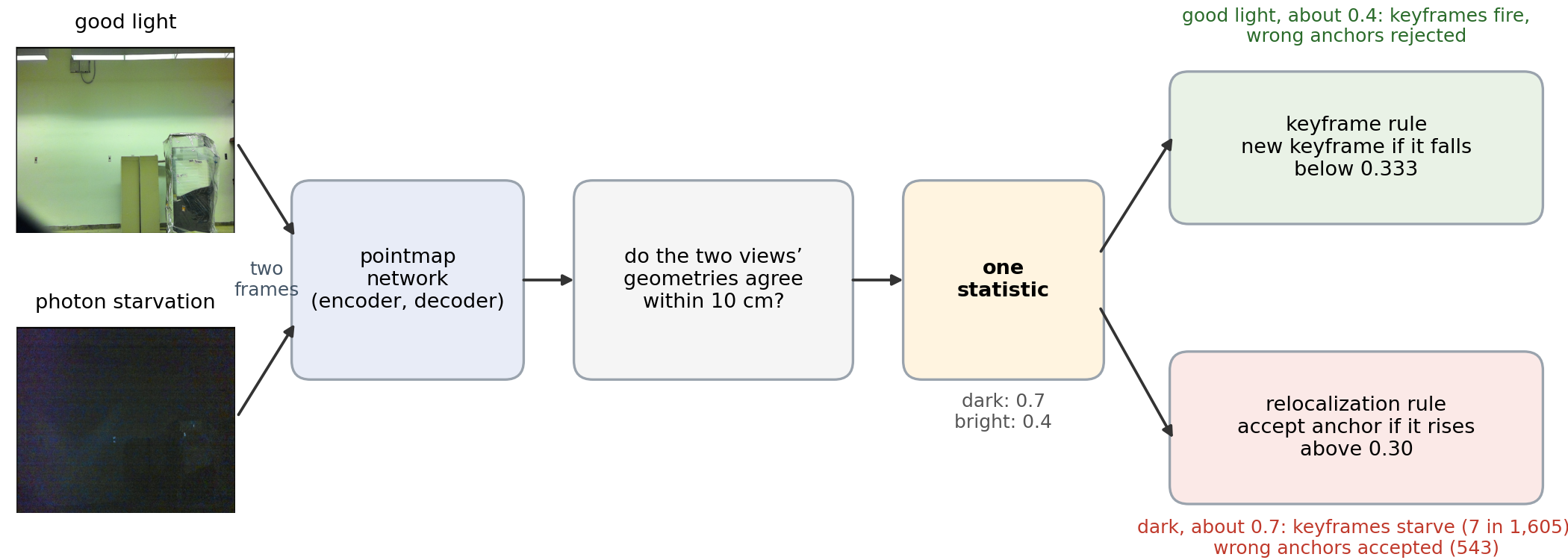}
\caption{The mechanism. Two frames enter the pointmap network; a correspondence is accepted when the two views' geometry predictions agree within 10~cm; the confident fraction is the one statistic both rules read. The keyframe rule fires when it falls below 0.333 and the relocalization rule accepts an anchor when it rises above 0.30. In good light it sits near 0.4 and both rules behave. Starved of structure, the network invents geometry that agrees with itself, the statistic rises to about 0.7, keyframing starves (7 keyframes in 1\,605 frames) and relocalization accepts wrong anchors (497 of 543 on \texttt{plant\_dark}). Frames: the robot's gripper camera, lights on and at 1/100~s.}
\label{fig:mech}
\vspace{-3.5mm}
\end{figure*}

Three sequences and two opposite symptoms converge on $m\approx0.7$ against thresholds near 0.3 (Fig.~\ref{fig:mech}): median post-gate $m$ 0.728 and 0.691 on the silent \texttt{sofa\_dark\_1} and \texttt{sofa\_dark\_3} at $g$ 0.929 and 0.927, 0.699 on \texttt{plant\_dark}'s relocalization edges at $g$ 0.831 (0.826 over all its calls), against 0.072 at $g$ 0.562 on the working \texttt{einstein\_dark}. The adapted model closes both exits by the same route: its post-gate $m$ on \texttt{sofa\_dark\_1}'s keyframe-pair edges is 0.219, below the threshold, and its relocalization-edge $m$ on \texttt{plant\_dark} is 0.0065, so 97.5\% of candidates are rejected. Intervening on one threshold at a time confirms the split: only the keyframe threshold moves the starvation symptom (7 to 49 keyframes), only the relocalization threshold moves the thrashing.

\begin{figure}[t]
\centering
\includegraphics[width=\columnwidth]{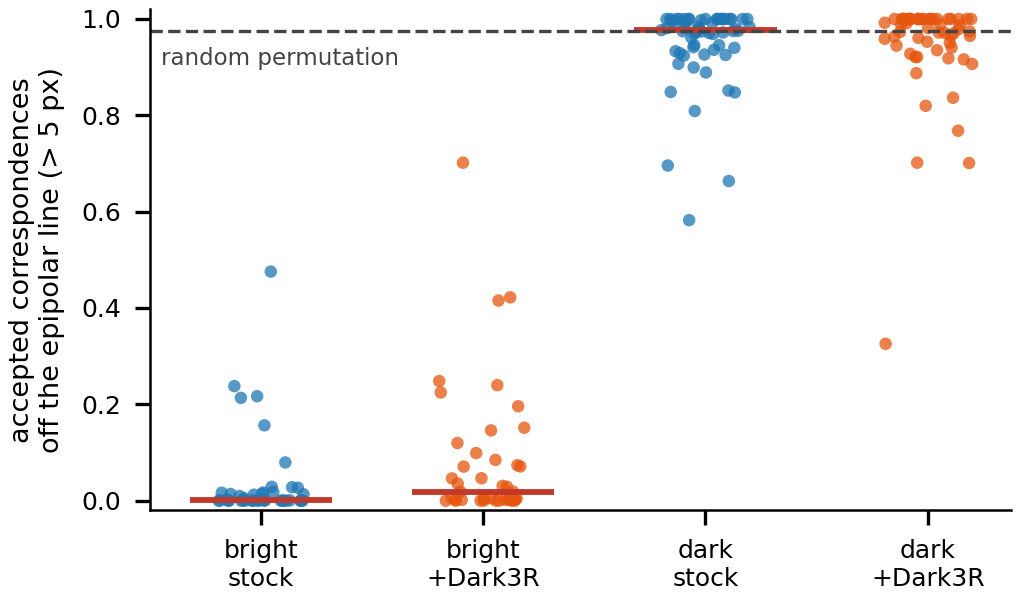}
\caption{The agreement is wrong. Per frame pair at a 40-frame baseline, the fraction of the correspondences each arm accepts (valid and above the confidence gate) that lie more than 5~px from the ground-truth epipolar line; red bars are medians; the dashed line is a random permutation of the same pixels. Bright pairs from two ETH3D sequences (42 per arm), dark pairs from three of the five dark sequences (61 per arm).}
\label{fig:epi}
\vspace{-3.5mm}
\end{figure}

\textbf{Is the agreement wrong? The epipolar test} (Fig.~\ref{fig:epi}). ETH3D ships calibration and poses, so the correspondences each arm accepts can be scored against the true epipolar geometry. On identical pairs from five dark and two bright sequences (172 dark pairs over two baselines; we report the 85 at 40 frames, since a 10-frame baseline is short enough to be self-consistent for any model), stock's accepted correspondences are 97--100\% more than 5~px from the epipolar line in the dark against 0.0--1.5\% in bright, and the adapted arm's 96--99\% against 0.2--7.9\%; a random permutation of the same pixels scores 97--98\%. Stock accepts 58--91\% of all pixels in the dark, the adapted arm 3--65\%. So stock's manufactured agreement is wrong agreement; the adapted arm's accepted correspondences are also wrong, so what un-starves keyframing is lowered confidence in wrong correspondence, not restored correct correspondence; and $g$ does not certify geometry: \texttt{merged/sofa\_dark\_1} reads 0.371 while 97\% of its accepted correspondences are wrong and its ATE is 1.32~m on a $\sim$3$\times$4~m room.

\textbf{The gate's own operand, every frame.} Logging the tracker's keyframe test on 36 replays (every original keyframe count reproduced) shows the starvation directly: on all six silent stock runs (\texttt{sofa\_dark\_1/2/3} at $\tau_Q$ 1.5 and 1.0) $\min(m_k,u_f)$ is at or above 0.333 on 99.3--100\% of frames, medians 0.85--0.89, and the rule fires 0--6 times in 277--1\,604 tracked frames. The same replays return a negative: the per-frame validity $g_{\mathrm{short}}$ reads 0.92--0.93 on silent runs but 0.88--0.91 on the working \texttt{kidnap\_dark} control; two views a few frames apart are self-consistent for any model, so the detector is the keyframe-pair statistic, whose evidence arrives only at factor-graph calls, 2--6 per silent run. The failure suppresses its own monitoring evidence; the first reading waits for the second keyframe, which motion sets (a bright stock run also waits 279 frames): frame 286, 150 and 231 (10.5, 5.6 and 8.5~s into sequences of 59, 44 and 10~s) on the three shipped-configuration silent runs, and never on the two one-keyframe stock runs at $\tau_Q=1.0$. Once a reading exists it is already diagnostic: on \texttt{sofa\_dark\_1} every one of stock's six readings sits near 0.93.

\begin{figure}[t]
\centering
\includegraphics[width=0.75\columnwidth]{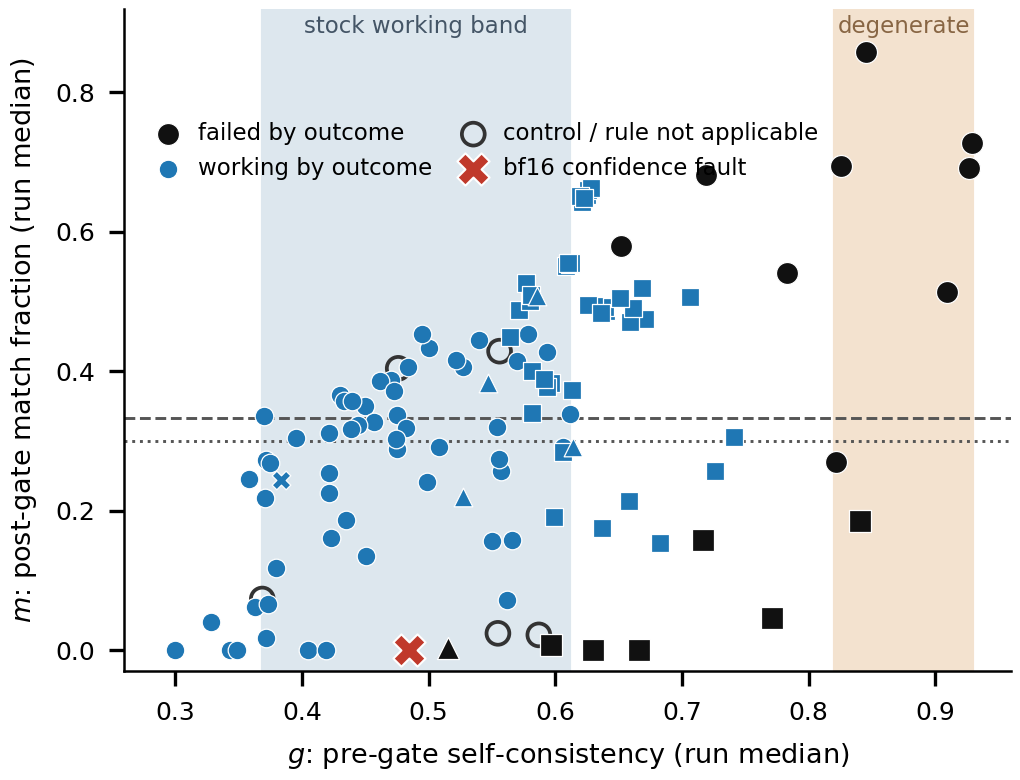}
\caption{The $(g,m)$ plane, one dot per run for 121 runs with a reading. Circles ETH3D 8-bit, squares Dark3R raw, triangles kitchen re-encoded to 8-bit; tints are the stock working band (0.369--0.611) and the degenerate band (0.821--0.929); dashed and dotted lines are the keyframe rule (0.333) and the relocalization rule (0.30) on $m$. Labels never use $g$ or $m$: an ETH3D dark run is working at $\geq 50$ keyframes with a converging alignment, thrashing at $\geq 100$ relocalization attempts, silent otherwise; a bright run is degenerate at $\geq 10$ Cholesky failures; a raw run is lost if it relocalized at all. Hollow: the \texttt{kidnap\_dark} controls and the two partial-dose \texttt{sofa\_dark\_2} runs.}
\label{fig:gmplane}
\vspace{-3.5mm}
\end{figure}

\textbf{Detector.} Across the 24-run grid (stock at $\tau_Q\in\{1.0,1.5\}$, adapted at $\{1.5,3.0\}$, 20 readings; four one-keyframe runs emit none, and zero factor-graph calls over a sequence of motion is itself the alarm, checked before any statistic), $g$ separates with no overlap: stock degenerate 0.821--0.929 ($n=5$), stock working 0.369--0.611 ($n=5$), adapted 0.328--0.451 ($n=10$). \texttt{kidnap\_dark} (0.369) fails without triggering it, correctly: a scripted kidnap is not degeneracy. \textbf{Labels never come from $g$}, and keyframe count and $g$ are decoupled in both directions: \texttt{plant\_dark} thrashes at 506 keyframes and $g$ 0.826, the robot's stock run at L3 holds 58 keyframes at 0.86--0.88, and bright sequences keep 14--26 keyframes per 1\,000 frames at $g$ 0.36--0.53. \textbf{Error rates and an operating point.} Over the 66 ETH3D runs labeled by outcome (11 failed, 55 working; three kidnap controls excluded), the area under the ROC is 0.95; at a threshold of 0.65, 9 of the 11 failures and none of the 55 working runs exceed it. Both numbers are in-sample (the 66 include the 20 grid runs and the threshold is chosen on them), and the two misses are the partial-dose runs on \texttt{sofa\_dark\_2} (4 and 41 poses in 281 frames), labeled failed by the $\geq 50$-keyframe rule at the limit of its reach. \textbf{What a deployer should use.} No transferable threshold exists yet. The absolute 0.65 false-alarms on a healthy robot walk (0.72, Sec.~\ref{sec:robot}); a ratio to the bright pass overlaps (working runs reach 1.63$\times$, the shipped failures start at 1.65$\times$). What transfers is the ordering: within one camera and one acceptance radius, a rise of $g$ above that camera's healthy runs marks manufactured agreement (it did on the two darkest robot routes and not on the two brighter, where no failure, loud or quiet, left an ordering); the loud exit is read from the relocalization count and post-gate $m$, and on a robot the map is checked against odometry, which needs no calibration. \textbf{It is graded}: along $W(s)$ on three dark sequences, $g$ falls on eight of eight steps read at both ends (\texttt{sofa\_dark\_1} 0.929, 0.783, 0.578, 0.371 at $s=0,0.5,0.75,1$ as keyframes rise 7, 18, 102, 201) and on five of six dark sequences, the exception being the \texttt{kidnap\_dark} control. \textbf{It needs no ground truth}: \texttt{kidnap\_2} ships none, and $g$ identified the degenerate arm (stock, 0.719, 78 Cholesky failures; the other arm 0.370). The one bright degeneracy, \texttt{merged/kidnap\_1} at 0.845, reverses on \texttt{kidnap\_2}: a kidnap destabilizes either arm; three shake sequences show no motion effect (both arms 0.554--0.606).

\textbf{On raw 14-bit input the failure is a different fault, and $g$ does not see it.} On two Dark3R raw scenes at nine brackets and five doses (45 runs) the stock model loses tracking at the dark brackets \emph{loudly}, 306--358 relocalization attempts per run, with post-gate $m$ collapsing to 0.000--0.159 against 0.154--0.663 for every run that tracks, while $g$ overlaps (lost 0.60--0.77, complete 0.56--0.74). \textbf{The encoding decides the kind of failure}: when the kitchen brackets are re-encoded to the 8-bit sRGB stream a camera would deliver, stock's failure at 1/1000 turns from loud to silent (360 frames, one pose, zero relocalizations, zero factor-graph edges, 11.6~FPS) while the sRGB shadow lift lets it track at 1/400, where linear raw lost it. The adapted model shows the mirror image: it tracks 1/1000 raw and thrashes on the 8-bit encoding of the same frames (303 relocalizations). Six runs on one scene, consistent with the noise-pair result of Sec.~\ref{sec:ablate}. \textbf{Two blind spots}: on a synthetic exposure ladder (three bright sequences, photon scaling with Poisson and read noise) adaptation extends tracking by about 7~dB but $g$'s bands blur to 0.55--0.75, and it does not see a fault that destroys confidence ordering but leaves geometry intact: bfloat16 inference reads 0.484, healthy, on a run with 94\% of edges rejected, which the post-gate $m$ (0.0001 against 0.245) catches. Fig.~\ref{fig:gmplane} places 121 runs on one plane: in the 8-bit ETH3D family every silent failure but the two \texttt{sofa\_dark\_2} dose runs sits right of every working run, and $m$ has no global threshold.

\subsection{Ablations: acceptance radius, no-scene inputs, partial merges}\label{sec:ablate}

\begin{figure*}[t]
\centering
\includegraphics[width=0.75\textwidth]{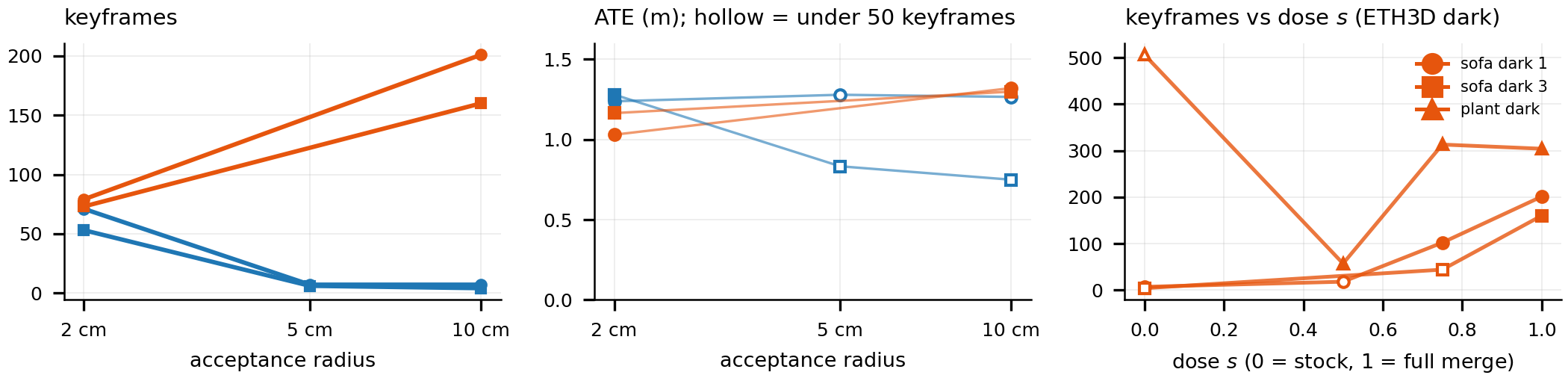}
\caption{Ablations. Left and middle: acceptance radius on \texttt{sofa\_dark\_1} (circles) and \texttt{sofa\_dark\_3} (squares), stock (blue) and adapted (orange): at 2~cm stock's keyframe count turns healthy while its ATE does not move; hollow markers mark runs under 50 keyframes, where ATE is not a measurement. Right: keyframes against dose $s$ on three ETH3D dark sequences; hollow markers are runs that starved or thrashed.}
\label{fig:ablate}
\vspace{-3.5mm}
\end{figure*}

\textbf{Tightening the acceptance radius hides the failure; it does not fix it} (Fig.~\ref{fig:ablate}). Halving \texttt{dist\_thresh} to 5~cm leaves the silent exit intact (\texttt{sofa\_dark\_1} 7 keyframes at $g$ 0.86, \texttt{sofa\_dark\_3} 6 at 0.85). At 2~cm the counts turn healthy: 71 and 53 keyframes, $g$ 0.46 and 0.50, 4 and 0 relocalizations. The maps are still wrong, ATE 1.24 and 1.28~m in a room of about 3$\times$4~m. The radius decides which correspondences pass the self-consistency test, not whether they are correct, so a run that would pass every count-based check still has the same wrong geometry. At 2~cm stock reaches \emph{usable} on four of the five sequences that completed, a third retune with the same wrong correspondences, while the adapted model loses tracking or overflows the keyframe store on 4 of 6 sequences at both 5 and 2~cm: the shipped 10~cm is where stock's manufactured agreement passes and the adapted model's sparser agreement still does, and tightening it removes the remedy. Because $g$ is defined through the radius, its bands hold at one radius only.

\textbf{The agreement is manufactured from the noise itself.} Two independent frames of 9-level quantized noise, with no scene at all, give stock a cross-view self-consistency of 0.936; the real dark pair gives 0.935. The adapted model reads 0.104 on the same noise. On the noise pair the confidence gate rejects everything; the real frames supply the confidence. The adapted model has a blind input of its own (uniform white noise, 0.976): each network self-agrees on some structureless input, and the 9-level input, shaped like ETH3D dark, is stock's.

\textbf{Neither half of the merge is the adaptation.} The encoder blocks alone un-starve keyframing on \texttt{sofa\_dark\_1} (7 to 68 keyframes, 44 relocalizations) while $g$ stays at 0.79, above the stock working band (0.369--0.611); the decoder blocks alone cannot track the sequence at all. Only the full merge moves both keyframing and $g$.

\subsection{Negative result: enhancement}\label{sec:enh}

DarkIR-m applied offline (a 31$\times$ brightness lift) and then stock MASt3R-SLAM, swept over four values of $\tau_Q$, yields \textbf{0/6} usable trajectories against stock's 2/6 at its own best setting: worse than no preprocessing, with at most 13 keyframes anywhere. Enhanced frames read $g=0.887$--$0.948$, the highest values we record, on every reading but the \texttt{kidnap\_dark} control (0.600): brightening frames that hold about nine intensity levels amplifies quantization into plausible gradients and manufactures the degeneracy signature.

\subsection{Embedded deployment}\label{sec:embedded}

On Jetson AGX Orin live CUDA allocation spans only 6.91--9.12~GB over runs of 4--394 keyframes while the allocator's reserve spans 7.3--54.3~GB: the reserve is an artifact, and the preallocated keyframe store gives a law, live demand $\approx 3.57$~GB $+\,6.54$~MB $\times$ \texttt{keyframe\_buffer} $+$ up to $\sim$2.2~GB of map growth, with an emulated Orin NX ceiling near 128 keyframes on working maps. Power is flat at 36.5--39.3~W across both arms; energy per keyframe is 6.2--17.2$\times$ lower for the adapted model over six pairs because stock produces 4--10 keyframes at the same power, while its energy per frame is 1.8--2.1$\times$ stock's, a keyframe-count effect that vanishes on the robot's ladder, where both arms keyframe (Sec.~\ref{sec:robot}).

\subsection{Negative results, and the dose}\label{sec:negatives}

Measured negative: bfloat16 inference loses dark tracking on both datasets (chapel 1 of 500 frames tracked, 246 relocalization attempts; ETH3D 1\,319 attempts) because an 8-bit mantissa destroys the confidence ordering the gate depends on; fp16 keeps it (500 of 500).

\textbf{The daylight tax and the dose.} On a 500-frame per-frame trajectory of a raw bright scene the adapted checkpoint's ATE is 0.421 against stock's 0.178 ($+137\%$; one scene, one exposure), where Dark3R's own clean-image table reports $+18\%$ offline~\cite{dark3r2026} and DarkVGGT names the same tax~\cite{darkvggt2026}; the figure rises monotonically with the dose $s$. Swept over all five Dark3R raw test scenes at four exposures (per-frame ATE against COLMAP), three regimes appear: in good light the full merge is never best; in moderate darkness a partial dose beats the full merge on ATE in 8 of 8 pairs; in extreme darkness the full merge is best on 3 of 3 and partial doses lose tracking. Both ends hold on our own protocol: ETH3D dark is the extreme regime (\texttt{sofa\_dark\_1} 7, 18, 102, 201 keyframes at $s=0,0.5,0.75,1$), and a partial dose removes \texttt{merged/kidnap\_1}'s bright degeneracy entirely ($g$ 0.439 at $s=0.5$ against 0.845, Cholesky failures 46 to 0). Every stock-on-linear-raw comparison carries an encoding confound (stock was trained on sRGB, whose shadow lift alone lets it track a kitchen bracket it loses on linear raw). A rule that raises $s$ until $g$ enters the healthy band is conceivable on ETH3D; we did not run it, and on the raw scenes it could not work, since $g$ does not separate lost from complete runs there. \textbf{The healthy level is a property of the camera and pipeline} (ETH3D working stock 0.369--0.611, raw chapel complete 0.60--0.74, raw kitchen 0.56--0.68): a deployment calibrates it on a bright pass of the same camera, reads $g$ per caller, checks tracking (zero factor-graph calls, the relocalization count, post-gate $m$) before any statistic, and on a robot checks motion against odometry: coverage flags each lost turn-free walk, the end-to-start extent the squashed maps (Sec.~\ref{sec:robot}).

\textbf{Acting on the alarm.} A wrapper reads $g$ at every factor-graph call and, the first time it exceeds 0.65, switches the live model to the adapted weights in place, once. On \texttt{sofa\_dark\_1} the alarm trips at the first reading (frame 286, $g$ 0.927) and the run goes from 7 keyframes to 187 (35 relocalization attempts, $g$ 0.399 after the switch, ATE 1.27~m over 187 poses; the adapted arm: 201 keyframes, 1.32~m); a half dose yields 13 keyframes. On the robot's L3 walk the alarm trips at frame 179 ($g$ 0.946) and the switched run reproduces the adapted arm's own failure (47 keyframes to its 49, 459 attempts, $m$ 0.000, $g$ 0.453): the action works where the adapted weights work and inherits their scope; the switch itself caused no loss (at most one Cholesky failure). Single runs, one sequence and one walk.

\section{Scope test on a robot}\label{sec:robot}

\begin{figure}[t]
\centering
\includegraphics[width=\columnwidth]{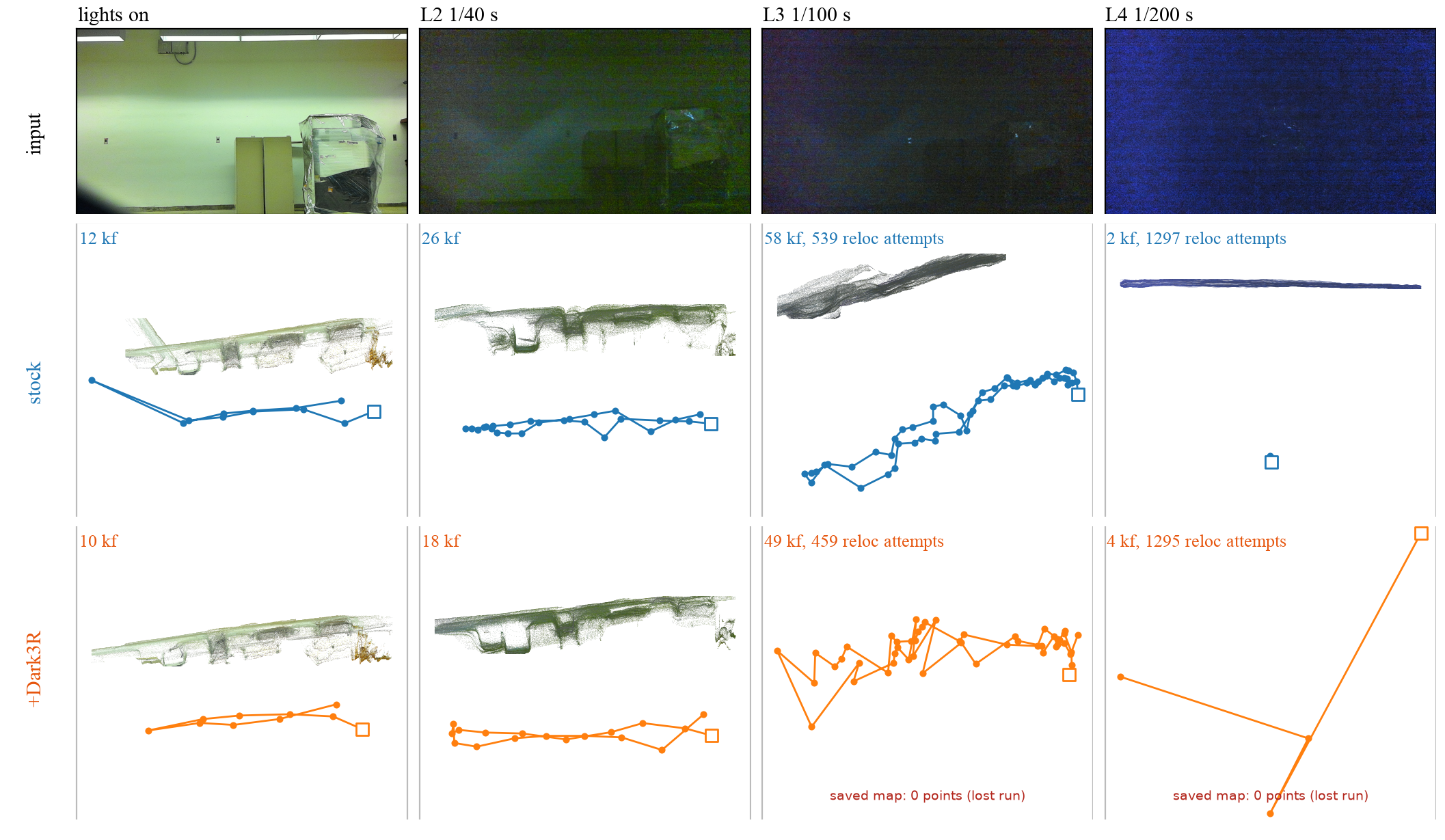}
\caption{The robot's ladder, first room. Top: four of the five levels as captured by the gripper camera (lights on 1/100~s gain 0.05; L2 1/40~s, L3 1/100~s, L4 1/200~s at gain 0.40). Below: the saved reconstruction and keyframe poses of each replay on the payload AGX Orin, stock and adapted, each run in its own scale (square: start). Both arms map the route through L2; at L3 both lose it and relocalize hundreds of times; at L4 neither maps the route.}
\label{fig:spot}
\vspace{-3.5mm}
\end{figure}

We treat the robot as a scope test, not a deployment: does each ETH3D finding reproduce on an 8-bit camera on hardware, and what governs whether it does? The whole stack (both weight sets, in-process backend, windowed factor graph, detector logging) runs on a Jetson AGX Orin carried as a quadruped's payload (a Boston Dynamics Spot), replaying walks recorded by its gripper camera (Fig.~\ref{fig:spot}). The camera delivers an 8-bit ISP-processed stream (no camera on the platform exposes raw); its controls are unitless. Automatic exposure, focus, HDR and white balance are off, so that the light level is a controlled variable per walk; a comparison against the camera's own auto-exposure is future work. Frames are stored uncompressed at 10~Hz; the reference trajectory is the robot's odometry at each image's acquisition time, a \emph{reference}, not ground truth (about 1\% drift).

\textbf{Protocol.} One room, one taped out-and-back route of about 24~m, 1\,190--1\,300 frames per walk. Lights on (walked twice), then lights off at four exposure levels with gain raised to hold the frame median near 40: 1/13~s (L1), 1/40~s (L2), 1/100~s (L3, walked twice), 1/200~s (L4). Each walk was replayed on the payload board with both weight sets (uncalibrated, window 120, 128-slot store), logging keyframes, relocalizations, $m$, $g$ per caller, FPS and power; a run is judged first by whether it tracked, then read on $g$.

\begin{table}[t]
\caption{Robot ladder, both arms, payload AGX Orin. One walk per level (repeats gave the same verdicts). kf path: odometry distance spanned by keyframes / walked. $g$ per caller (backend / reloc). J: energy per frame.}
\label{tab:robot}
\centering\scriptsize
\setlength{\tabcolsep}{2pt}\begin{tabular}{llrlrlrr}
\toprule
level & arm & kf & reloc ok/att & $m$ & $g$ & kf path (m) & J \\
\midrule
lights on & stock & 12 & 0/0 & 0.36 & 0.49 & 22.0/25.0 & 13.7 \\
 & +Dark3R & 10 & 0/0 & 0.41 & 0.41 & 19.6/25.0 & 13.3 \\
L1 1/13\,s & stock & 17 & 0/0 & 0.37 & 0.49 & 21.6/23.3 & 13.6 \\
 & +Dark3R & 10 & 0/0 & 0.36 & 0.44 & 19.8/23.3 & 13.4 \\
L2 1/40\,s & stock & 26 & 0/0 & 0.29 & \textbf{0.72} & 21.7/22.7 & 14.0 \\
 & +Dark3R & 18 & 0/0 & 0.32 & 0.61 & 21.7/22.7 & 13.8 \\
L3 1/100\,s & stock & 58 & 2/539 & \textbf{0.07} & \textbf{0.86/0.88} & 19.1/25.0 & 33 \\
 & +Dark3R & 49 & 1/459 & \textbf{0.000} & 0.45/0.49 & \textbf{11.7}/25.0 & 31 \\
L4 1/200\,s & stock & \textbf{2} & 1/1297 & 0.000 & --/0.73 & 0/23.9 & 41 \\
 & +Dark3R & \textbf{4} & 0/1295 & 0.000 & 0.80/0.76 & 0/23.9 & 56 \\
\bottomrule
\end{tabular}
\vspace{-2mm}
\end{table}

\begin{figure*}[t]
\centering
\includegraphics[width=0.75\textwidth]{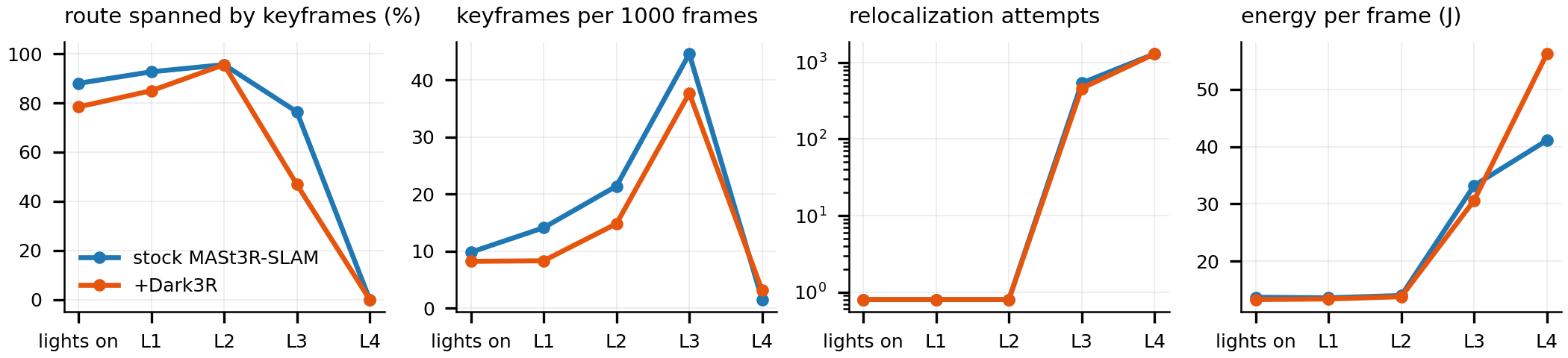}
\caption{The robot's ladder in numbers (Table~\ref{tab:robot}), both arms, five light levels: route spanned by keyframes (keyframe path over path walked), keyframes per 1\,000 frames, relocalization attempts (log scale; zero drawn at the floor) and energy per frame from the full power trace. Nothing separates the arms until L3, where both fail loudly; the adapted arm's keyframes span half the route, and the loud failure costs 2--4$\times$ the energy per frame.}
\label{fig:ladder}
\vspace{-3.5mm}
\end{figure*}

\textbf{What the robot shows} (Table~\ref{tab:robot}, Figs.~\ref{fig:spot} and \ref{fig:ladder}). \emph{Both arms track through L2, and $g$ grades the darkness for both}: no relocalization on either arm, $g$ rising while tracking holds (stock 0.49, 0.49, 0.72; adapted 0.41, 0.44, 0.61). Stock's healthy L2 walk reads 0.72, above the 0.65 operating point, a false alarm on a healthy run: the level is calibrated per camera. \emph{At L3 both arms lose tracking, loudly}: stock relocalizes 539 times with 2 successes and its $m$ collapses to 0.07; the adapted arm 459 times, $m$ 0.000, covering half the route. At L4 both are lost from the first frames and the adapted arm's $g$ (0.80 / 0.76) also rises past ETH3D's working band: 1/200~s degenerates either model. \emph{On the same loud failure, $g$ reads stock's collapse and the adapted arm's non-collapse}: at L3 stock's $g$ is 0.86 / 0.88, inside ETH3D's degenerate band on a camera used to build no band, the adapted arm's 0.40--0.49; the adapted weights remove the self-agreement even here and the arm fails by confidence instead. \emph{On this ladder the remedy does not appear}: no level exists at which stock fails and the adapted arm tracks, nor at a half dose, which tracks the same three levels (a brief loss at L1), loses the same two, and at L3 overflows the 128-slot store with relocalization-appended keyframes. All fourteen replays repeat on the desktop with identical keyframe counts and $g$ within 0.04.

\begin{table}[t]
\caption{Three routes in brighter rooms that never revisit their start, both arms, 512-slot store. Coverage: odometry path spanned by keyframes over path walked. Lit controls cover 93--97\% on both arms. Structure SNR at 1~ms: $+4.7$, $-0.3$ and $+9.1$~dB (first room's failing level: $-4.6$).}
\label{tab:routes3}
\centering\scriptsize
\setlength{\tabcolsep}{2pt}\begin{tabular}{llrlrrlr}
\toprule
route & shutter & \multicolumn{3}{c}{stock} & \multicolumn{3}{c}{+Dark3R} \\
 & & kf/1k & reloc & cov.\% & kf/1k & reloc & cov.\% \\
\midrule
L-shape, 12.5~m, along & 2.5~ms & 13 & 0/359 & \textbf{41} & 123 & 0/0 & \textbf{100} \\
 & 1~ms & 110 & 0/0 & 100 & 152 & 0/0 & 100 \\
strafe, 6.7~m, across & 2.5~ms & 538 & 0/129 & \textbf{48} & 298 & 0/0 & \textbf{100} \\
 & 1~ms & 28 & 2/298 & 0 & 90 & 0/246 & 25 \\
strafe, 9.4~m, across & 2.5~ms & 238 & 0/0 & 100 & 73 & 0/0 & 95 \\
 & 1~ms & 120 & 0/281 & 0 & 30 & 0/387 & 0 \\
\bottomrule
\end{tabular}
\vspace{-2mm}
\end{table}

\textbf{Three routes that never revisit their start} (Table~\ref{tab:routes3}). Three routes in brighter rooms were walked lit and at 2.5 and 1~ms shutter at gain 0.40; 1~ms is the camera's floor, and the stationary-start structure SNR there was $+4.7$, $-0.3$ and $+9.1$~dB against $-4.6$~dB at the first room's failing level, so only one of these routes approaches that darkness. One route is L-shaped with the camera facing along the walk; two are strafes with the camera across it. Every lit control reconstructs. Three things follow. \emph{At 2.5~ms the adaptation covers a level stock loses}, on two routes of three: stock keeps 41\% and 48\% of the route with 359 and 129 failed relocalization attempts, the adapted arm keeps all of it; on the third both track. \emph{At 1~ms both arms fail loudly on both strafes} (0--25\% coverage, 246--387 attempts, $m$ 0.000, 98--100\% of edges rejected, 45--57~J per frame against 14--15 lit). On three of those four failed runs every keyframe was created during the stationary start and tracking failed once the robot moved, so their backend statistics describe a stationary camera. \emph{On the L-shape at 1~ms both fail quietly, in a different shape}: zero attempts, $g$ at its lit level, stock at five times its lit keyframe rate, and a complete-looking map whose end-to-start extent, relative to the odometry's, falls as exposure and gain drop: 0.86--0.87 lit, 0.55 at 2.5~ms (the adapted 100\% map), 0.43--0.48 at 1~ms, 0.38 at low gain. This is the quiet face on the one route where the camera looks along the walk, the direction a single camera observes least well. $g$ orders only on the darkest of the three routes: on the $-0.3$~dB strafe stock's 1~ms collapse reads 0.556 against its lit 0.251 (a stationary-start run; the moving adapted run there 0.387 against 0.257), while on the brighter two every failed run reads 0.17--0.36 against lit controls of 0.32--0.36.

\textbf{Cost, and what we do not claim.} While tracking the two arms cost the same, 13.2--13.9~J per frame at 34--35~W and 2.6~FPS against the 10~Hz recording (Table~\ref{tab:robot}); the loud failure costs 22--34~J per frame at L3 and 41--56 at L4, 1.6--4$\times$ the tracking cost, at 0.75--1.7~FPS. ATE against odometry (0.24--0.73~m over 10--26 poses) is not a result. Four rooms, four routes, eighteen walks, every row a single run; accuracy is not measured.

\section{Conclusion and limitations}\label{sec:concl}

Dense visual SLAM built on learned two-view geometry fails in low light by one mechanism: spurious self-consistency of the network's cross-view geometry inflates a match statistic past thresholds that two subsystems read in opposite directions, producing a near-empty map reported as healthy or unbounded relocalization onto wrong anchors. The quantity responsible is itself a detector: no computation, no ground truth. Low-light adaptation repairs both symptoms on ETH3D by withdrawing confidence, not by finding right geometry; its good-light cost scales with the dose, and the detector reads the cure and the good-light degeneracy, not the ATE tax. On a quadruped's 8-bit camera the mechanism's signature reproduces on the two darkest routes of four, the adaptation covers one dark level on two of three turn-free routes, and which face of the fault appears varies with route geometry and sensor encoding.

\textbf{Limitations, each stated once.} $g$ does not certify geometry: a working-band reading means the network is not degenerately self-consistent, not that the map is right (\texttt{merged/sofa\_dark\_1}: 0.371, 97\% wrong). \textbf{Evidence arrives only at keyframes:} 2--6 readings on a silent run, the first after 5.6--10.5~s; the per-frame counterpart is dominated by short baselines; a scheduled wide-baseline probe was not built. \textbf{Scope.} $g$'s evidence is the 8-bit ETH3D silent-failure regime: it does not transfer to synthetic noise, does not see confidence collapse (bfloat16, raw input, the robot) or squashed maps, its healthy level is calibrated per camera, and its operating point is in-sample. \textbf{One system; the alarm acted on once.} Whether other pointmap SLAM systems share the mechanism is untested; no feature-based or inertial baseline was run. Two bright textureless sequences read $g$ 0.31--0.34 on both arms: low texture does not inflate self-consistency; pure rotation is untested. The alarm-triggered switch of Sec.~\ref{sec:negatives} is one run per setting and switches once, forward; light was constant within every run, so recovery when it returns is unmeasured. \textbf{Other limits.} One enhancer and one adaptation were tested; energy ratios are ranges (determinism does not survive a power-mode change); odometry is a reference, no depth was recorded. Code and the merge recipe will be released; the merged checkpoint cannot be (CC BY-NC-SA base weights).

\section*{Acknowledgment}
An AI assistant (Claude Opus 5) was used for text drafting and editing and for the scripts that generate Figs.~1--7; the authors reviewed and verified all such output. The ideas, implementation, experiments, measurements, analysis, and conclusions are the authors' own.

\vspace{-1.5mm}
\bibliographystyle{IEEEtran}
{\footnotesize \bibliography{references}}

\begin{thebibliography}{10}
\providecommand{\url}[1]{#1}
\csname url@samestyle\endcsname
\providecommand{\newblock}{\relax}
\providecommand{\bibinfo}[2]{#2}
\providecommand{\BIBentrySTDinterwordspacing}{\spaceskip=0pt\relax}
\providecommand{\BIBentryALTinterwordstretchfactor}{4}
\providecommand{\BIBentryALTinterwordspacing}{\spaceskip=\fontdimen2\font plus
\BIBentryALTinterwordstretchfactor\fontdimen3\font minus
  \fontdimen4\font\relax}
\providecommand{\BIBforeignlanguage}[2]{{%
\expandafter\ifx\csname l@#1\endcsname\relax
\typeout{** WARNING: IEEEtran.bst: No hyphenation pattern has been}%
\typeout{** loaded for the language `#1'. Using the pattern for}%
\typeout{** the default language instead.}%
\else
\language=\csname l@#1\endcsname
\fi
#2}}
\providecommand{\BIBdecl}{\relax}
\BIBdecl

\bibitem{murai2025mast3rslam}
R.~Murai, E.~Dexheimer, and A.~J. Davison, ``{MASt3R-SLAM}: Real-time dense
  {SLAM} with {3D} reconstruction priors,'' in \emph{CVPR}, 2025.

\bibitem{ec3r2026}
L.~Hu \emph{et~al.}, ``{EC3R-SLAM}: Efficient and consistent monocular dense
  {SLAM} with feed-forward {3D} reconstruction,'' \emph{arXiv preprint
  arXiv:2510.02080}, 2025.

\bibitem{kumar2026uavslam}
P.~Kumar, A.~Deepak, and S.~Kumar, ``Robust visual {SLAM} for {UAV} navigation
  in {GPS}-denied and degraded environments: A multi-paradigm evaluation and
  deployment study,'' \emph{arXiv preprint arXiv:2605.03678}, 2026.

\bibitem{dark3r2026}
A.~Y. Guo \emph{et~al.}, ``{Dark3R}: Learning structure from motion in the
  dark,'' in \emph{CVPR}, 2026.

\bibitem{hu2022lora}
E.~J. Hu \emph{et~al.}, ``{LoRA}: Low-rank adaptation of large language
  models,'' in \emph{ICLR}, 2022.

\bibitem{schoeps2019eth3d}
T.~Sch{\"o}ps, T.~Sattler, and M.~Pollefeys, ``{BAD SLAM}: Bundle adjusted
  direct {RGB-D} {SLAM},'' in \emph{CVPR}, 2019.

\bibitem{wang2024dust3r}
S.~Wang \emph{et~al.}, ``{DUSt3R}: Geometric {3D} vision made easy,'' in
  \emph{CVPR}, 2024.

\bibitem{leroy2024mast3r}
V.~Leroy, Y.~Cabon, and J.~Revaud, ``Grounding image matching in {3D} with
  {MASt3R},'' in \emph{ECCV}, 2024.

\bibitem{wang2025vggt}
J.~Wang \emph{et~al.}, ``{VGGT}: Visual geometry grounded transformer,'' in
  \emph{CVPR}, 2025.

\bibitem{yang2025pi3}
Y.~Wang \emph{et~al.}, ``$\pi^3$: Permutation-equivariant visual geometry
  learning,'' \emph{arXiv preprint arXiv:2507.13347}, 2025.

\bibitem{maggio2025vggtslam}
D.~Maggio, H.~Lim, and L.~Carlone, ``{VGGT-SLAM}: Dense {RGB} {SLAM} optimized
  on the {SL}(4) manifold,'' in \emph{NeurIPS}, 2025.

\bibitem{thomas2026failuredrift}
A.~S. Thomas, S.~Agnihotri, and M.~Keuper, ``Failure or drift? {Evaluating}
  monocular {SLAM} under synthetic and real-world corruptions,'' in
  \emph{NeuSLAM Workshop at {ECCV}}, 2026.

\bibitem{basystyi2026lowlightslam}
O.~Basystyi \emph{et~al.}, ``{SLAM} in low-light environments: Project
  report,'' \emph{arXiv preprint arXiv:2607.17699}, 2026.

\bibitem{rabiee2020ivslam}
S.~Rabiee and J.~Biswas, ``{IV-SLAM}: Introspective vision for simultaneous
  localization and mapping,'' in \emph{CoRL}, 2020.

\bibitem{gaus2025super}
J.~A. Gaus, D.~H{\"a}ufle, and W.-J. Baek, ``{SUPER} -- {A} framework for
  sensitivity-based uncertainty-aware performance and risk assessment in visual
  inertial odometry,'' \emph{arXiv preprint arXiv:2512.14189}, 2025.

\bibitem{sansoni2025confidencetrap}
S.~Sansoni and S.~R. Tosetti~Sanz, ``The {SLAM} confidence trap,'' in
  \emph{IEEE Int. Conf. on Advanced Robotics ({ICAR})}, 2025, pp. 733--738.

\bibitem{xiong2026leangate}
X.~Xiong \emph{et~al.}, ``Accelerating transformer-based monocular {SLAM} via
  geometric utility scoring,'' \emph{arXiv preprint arXiv:2604.08718}, 2026.

\bibitem{darkvggt2026}
M.~Kweon \emph{et~al.}, ``{DarkVGGT}: Seeing through darkness using thermal
  geometry without daylight tax,'' \emph{arXiv preprint arXiv:2606.11326},
  2026.

\bibitem{twilightslam2023}
S.~P. Singh \emph{et~al.}, ``Twilight {SLAM}: Navigating low-light
  environments,'' \emph{arXiv preprint arXiv:2304.11310}, 2023.

\bibitem{darkir2025}
D.~Feijoo \emph{et~al.}, ``{DarkIR}: Robust low-light image restoration,'' in
  \emph{CVPR}, 2025.

\bibitem{kong2021nir}
D.~Kong, Y.~Zhang, and W.~Dai, ``Direct near-infrared-depth visual {SLAM} with
  active lighting,'' \emph{IEEE Robotics and Automation Letters}, vol.~6,
  no.~4, pp. 7057--7064, Oct. 2021.

\bibitem{clearir2026}
N.~Shankar, P.~Ladosz, and H.~Yin, ``{CLEAR-IR}: Clarity-en\-hanced active
  reconstruction of infrared imagery,'' \emph{arXiv preprint arXiv:2510.04883},
  2025.

\end{thebibliography}

\end{document}